\documentclass{IOS-Book-Article}

\usepackage{mathptmx}
\usepackage{amsmath,amssymb,bm}
\usepackage{graphicx}
\usepackage{enumitem}

\newcommand{\Sgeom}{S_{\mathrm{geom}}}
\newcommand{\Sreach}{S_{\mathrm{reach}}}
\newcommand{\Safford}{S_{\mathrm{afford}}}
\newcommand{\Savoid}{S_{\mathrm{avoid}}}
\newcommand{\Stask}{S_{\mathrm{task}}}

\newcommand{\Pobj}{\mathcal{P}_{\mathrm{obj}}}

\newcommand{\rev}[1]{#1}

\def\hb{\hbox to 11.5 cm{}}

\begin{document}

\pagestyle{headings}
\def\thepage{}

\begin{frontmatter}

\title{SAGP: Semantic Affordance-Guided Grasp Planning via Coarse-Zone VLM Reasoning}

\markboth{}{}

\author{\fnms{Muhayy} \snm{Ud Din}}
and
\author{\fnms{Irfan} \snm{Hussain}\thanks{Corresponding Author: Irfan Hussain, KU Center for Autonomous Robotic Systems (KUCARS), Khalifa University, United Arab Emirates; E-mail: irfan.hussain@ku.ac.ae.}}

\runningauthor{M. Ud Din and I. Hussain}
\address{KU Center for Autonomous Robotic Systems (KUCARS), Khalifa University, \\ United Arab Emirates}

\begin{abstract}
Geometry-based grasp planners ensure physically valid grasps but ignore
functional semantics, often generating grasps that are antipodal and
collision-free yet practically inappropriate, for example, gripping a mug by its
rim, a knife by the blade, or a bottle near its cap. These inconsistencies
cause the downstream task to fail even when traditional grasp metrics are met.
Existing vision-language model (VLM) approaches either depend on fine-grained,
category-specific part segmentation or attempt to directly infer grasp poses,
with the latter prone to spatial hallucinations. As a result, no practical,
training-free framework has yet been proposed that robustly links high-level
semantic reasoning to geometric grasp planning.
We introduce \textit{Semantic Affordance-Guided Grasp Planning (SAGP)}, a
training-free pipeline built on a \emph{coarse-zone abstraction layer}. The
method first partitions the object point cloud into spatial regions
(top, middle, bottom, lateral sides, and protrusions) by applying PCA-based
alignment followed by distance-driven DBSCAN clustering, entirely bypassing
learned segmentation. A pre-trained VLM then assesses the grasp quality of
each region through a structured zero-shot query, and the resulting zone-wise
scores are fused with geometric, reachability, and task-alignment signals to
re-rank antipodal grasp candidates. Experiments on YCB objects in PyBullet
with a Franka Panda robot show that SAGP preserves the high success rate of
geometry-only planning while substantially improving the
\emph{functional appropriateness} of selected grasps, particularly on
asymmetric, handle-bearing objects where geometry alone is uninformative.
The introduced coarse-zone abstraction offers an effective, training-free
bridge between VLM-based reasoning and geometric grasp planning, without the
need for fine-grained part segmentation.
\end{abstract}

\begin{keyword}
Grasp planning \sep Vision-language models \sep Semantic grasping \sep Robot manipulation
\end{keyword}
\end{frontmatter}

\section{Introduction}

The robotic grasping of everyday objects is fundamental for autonomous manipulation. Geometric planners, including force closure methods~\cite{bicchi2000}, GraspIt!~\cite{miller2004},
Grasp Pose Detection (GPD)~\cite{tenpas2017}, Dex-Net~\cite{mahler2017},
Contact-GraspNet~\cite{sundermeyer2021}, and AnyGrasp~\cite{fang2023} optimize
physical viability either by maximizing force-closure quality or by learning grasp
likelihood directly from point clouds.
These methods treat every object as a geometry to be grasped rather than an
artefact with functional structure.A purely geometric planner is just as likely to grasp a mug by the handle as by the rim, a knife by the handle as by the blade, or a drill by the grip as by the motor housing.
Humans naturally incorporate semantic understanding into grasping: a mug is
lifted by its handle, a bottle is held upright, and scissors are passed handle-first.
\rev{Human grasping also couples the choice of contact region with real-time
regulation of grip force in response to load, an ability that has inspired
human-inspired force-selection policies for robotic execution~\cite{stachowsky2015};
SAGP addresses the complementary problem of selecting the functionally
appropriate contact region.}
This gap between what is physically feasible and what is functionally suitable leads to grasps that satisfy force-closure criteria but still fail at the subsequent task.

Recent work has attempted to incorporate semantics or vision-language models (VLMs) into grasping, but existing approaches fall into three main categories, each with significant limitations.
First, methods that prompt VLMs to predict grasp poses or coordinates directly suffer from spatial hallucination, since these models are not designed for precise geometric reasoning~\cite{wake2023}.
Second, affordance-based approaches~\cite{murali2020,detry2011,tang2023,li2024} rely on dense part annotations or category-specific segmentation networks, which require training data for each object class and do not generalise well in a zero-shot setting.
Third, task-level systems such as SayCan~\cite{ahn2022}, VoxPoser~\cite{huang2023}, CLIPort~\cite{shridhar2022}, RT-2~\cite{brohan2023}, and Code-as-Policies~\cite{liang2023} operate at a higher level of abstraction and do not integrate directly with geometric grasp generation.
\rev{Closely related task-and-motion-planning tools couple an LLM-based symbolic planner with a geometric and kinodynamic motion planner~\cite{din2026lang2manip}, but reason at the level of symbolic actions and collision-free trajectories rather than selecting functionally appropriate grasp contacts.}
As a result, there is still no practical, zero-shot, training-free method that connects high-level semantic reasoning with low-level grasp planning.

Our key insight is that coarse spatial decomposition provides the appropriate interface between these two domains. Instead of relying on fine-grained segmentation or precise coordinate prediction, we partition an object into a small set of geometric zones such as top, middle, bottom, lateral regions, and protrusions. Vision-language models trained on large-scale image-text data are well suited to reason about such spatially described regions, including concepts like top, handle, blade, and grip, but remain unreliable at predicting exact coordinates or dense part labels. A discrete rating scheme consisting of good, acceptable, bad, and dangerous aligns with VLM confidence and provides sufficient structure to guide grasp selection.
Building on this insight, we propose SAGP (\textit{Semantic Affordance-Guided Grasp Planning}), a training-free pipeline that integrates zero-shot VLM reasoning with geometric grasp planning through a novel coarse-zone abstraction layer. The method decomposes point clouds using PCA-based alignment and DBSCAN-based clustering, queries a pre-trained VLM for zone-level affordance scores, and re-ranks antipodal grasp candidates using a unified multi-component objective.

We make three contributions:
\begin{enumerate}[leftmargin=1.5em, topsep=2pt, itemsep=1pt]
\item A \textit{coarse-zone abstraction layer} based on PCA axis alignment and
DBSCAN protrusion detection, which serves as a training-free interface
between VLM reasoning and geometric grasp planning, requiring no
category-specific segmentation or part annotations.
\item A \textit{training-free SAGP pipeline} that combines antipodal grasp
sampling, zero-shot VLM-derived zone affordances, and a fused five-component
scoring function balancing geometric, reachability, and semantic signals.
\item An \textit{evaluation on 14 YCB objects across three task types} in
PyBullet with a Franka Panda, including a per-category analysis on asymmetric
versus near-symmetric objects that isolates where semantic re-ranking adds
value.
\end{enumerate}

\section{Related Work}

\subsection{Geometric Grasp Planning}

Geometric grasp planning has been extensively studied, with early work such as force-closure analysis~\cite{bicchi2000} and GraspIt!~\cite{miller2004} establishing the theoretical foundations for multi-finger grasp synthesis.
Subsequent approaches have focused on scaling grasp generation and improving robustness from perception. GPD~\cite{tenpas2017} introduced antipodal grasp sampling directly from point clouds with learnt scoring, while Dex-Net~\cite{mahler2017} uses large-scale simulation to train grasp quality models on millions of samples. More recent methods such as Contact-GraspNet~\cite{sundermeyer2021} predict full 6-DOF grasp poses from depth observations using learned geometric representations, and AnyGrasp~\cite{fang2023} extends this capability to cluttered scenes through dense geometry matching.
\rev{Beyond contact synthesis, related work has addressed planning the grasping \emph{motions} of multi-fingered humanoid hands using low-dimensional synergy representations of the hand configuration space~\cite{rosell2019humanoid}.}

Despite their success, these methods share a common limitation: they operate purely on geometry and do not distinguish between functionally meaningful object regions. As a result, grasps that are physically valid may still be unsuitable for the intended task. SAGP addresses this gap by introducing a semantic re-ranking stage on top of standard antipodal sampling, preserving geometric robustness while incorporating functional preferences.

\subsection{Semantic and Task-Oriented Grasping}

To address the limitations of geometry-only methods, prior work has explored incorporating semantic and task-level information into grasping. TaskGrasp~\cite{murali2020} learns task-specific grasp strategies from a labelled dataset of 250 objects across 14 tasks using graph neural networks. Earlier work on grasp affordance densities~\cite{detry2011} models part-level contact distributions from visual demonstrations. More recent approaches such as GraspGPT~\cite{tang2023} use large language models to reason about grasp semantics, but require predefined part labels, which restricts zero-shot applicability. Similarly, SemGrasp~\cite{li2024} and LAN-grasp~\cite{mirjalili2023} extend semantic grasping to language-conditioned settings but depend on trained segmentation backbones and annotated data.

These approaches improve functional awareness but rely on either supervision or object-specific representations. In contrast, SAGP eliminates the need for part annotations and category-specific training. Its coarse-zone decomposition provides a generic proxy for object structure, enabling zero-shot generalisation to previously unseen objects.

\subsection{Vision-Language Models in Robotics}

Vision-language and language models have recently been applied to robotic reasoning and control, particularly at the task and planning level. SayCan~\cite{ahn2022} and VoxPoser~\cite{huang2023} use language models to compose sequences of manipulation skills, but do not address low-level grasp generation. CLIPort~\cite{shridhar2022} and RT-2~\cite{brohan2023} learn end-to-end visuomotor policies conditioned on language, requiring large-scale robotic datasets for training. Code-as-Policies~\cite{liang2023} translates language into executable robot programs, but relies on predefined primitives.

Recent work by Wake et al.~\cite{wake2023} highlights a key limitation of vision-language models in robotics, namely their unreliability in predicting precise spatial coordinates. SAGP builds directly on this observation by using VLMs only at a level where they are reliable. Instead of predicting grasp poses, the model provides zone-level categorical ratings, while geometric reasoning is handled entirely by the underlying grasp planner. This division of responsibility enables a practical and training-free integration of semantic reasoning into grasp planning.

\section{Methodology}

\begin{figure}[t]
  \centering
  \includegraphics[width=0.85\linewidth]{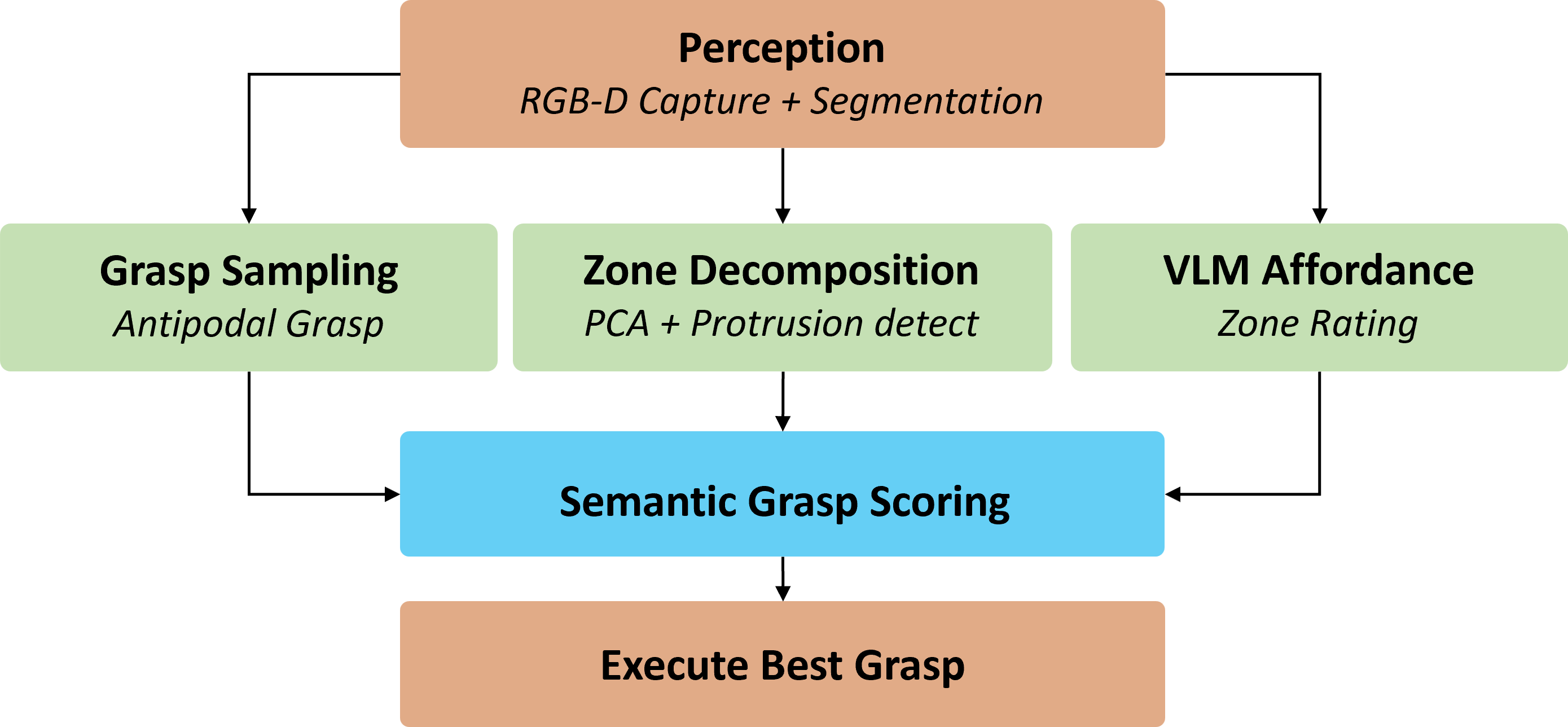}
  \caption{SAGP pipeline architecture. Starting from RGB-D perception
    and segmentation, the pipeline branches into three parallel
    processes: antipodal grasp sampling, coarse-zone decomposition
    via PCA and protrusion detection, and VLM-based zone affordance rating.
    The three streams converge at the semantic grasp scoring stage,
    where a fused five-component function re-ranks the candidates. The
    highest-ranked grasp is then executed.}
  \label{fig:pipeline}
\end{figure}

SAGP operates as a five-stage pipeline: (1) RGB-D perception and point cloud extraction, (2) antipodal grasp candidate generation, (3) coarse-zone decomposition, (4) VLM-based affordance extraction with structured constraints, and (5) fused semantic scoring and execution. The entire pipeline is training-free.
Figure~\ref{fig:pipeline} illustrates the overall architecture.
 
\subsection{Perception and Point Cloud Extraction}
 
We obtain aligned RGB, depth, and segmentation images from PyBullet using
\texttt{getCameraImage()} at a resolution of $640{\times}480$ with a
$60^\circ$ field of view. Raw depth-buffer values $z_{\mathrm{buf}}$ are
converted to metric depth via:
\begin{equation}
  d_i = \frac{f \cdot n}{f - (f - n)\, z_{\mathrm{buf},i}}
\end{equation}
where $f$ and $n$ denote the far and near clipping planes. Each pixel is then
back-projected into 3-D using the pinhole camera model:
\begin{equation}
  \mathbf{p}_i =
  \begin{bmatrix}
    (u_i - c_x)\, d_i / f_x \\
    (v_i - c_y)\, d_i / f_y \\
    d_i
  \end{bmatrix}
\end{equation}
The segmentation mask isolates the object point cloud $\Pobj$. To improve
surface coverage we capture views from one overhead and two side cameras and
merge them. Surface normals are estimated using Open3D~\cite{zhou2018} with a
radius of $0.02$\,m and 30 neighbours, oriented outward from the object
centroid. A tightly cropped RGB image with 20\% padding is also saved for the
VLM query.
 
\subsection{Antipodal Grasp Candidate Generation}
 
We generate grasp candidates by sampling $5{,}000$ point pairs from $\Pobj$
and filtering them with two standard geometric constraints:
\begin{itemize}[leftmargin=1.5em, topsep=1pt, itemsep=0pt]
  \item \textit{Gripper width}:
    $0.005 \le \|\mathbf{p}_i - \mathbf{p}_j\| \le 0.08$\,m;
  \item \textit{Antipodal condition}:
    $\angle(\mathbf{n}_i,\,\hat{\mathbf{a}}) < 45^\circ$ and
    $\angle(\mathbf{n}_j,\,-\hat{\mathbf{a}}) < 45^\circ$, with
    $\hat{\mathbf{a}} = (\mathbf{p}_j - \mathbf{p}_i)/\|\mathbf{p}_j-\mathbf{p}_i\|$.
\end{itemize}
The geometric quality of each surviving candidate is computed as:
\begin{equation}
  \Sgeom = \cos\!\bigl(\angle(\mathbf{n}_i,\,\hat{\mathbf{a}})\bigr)\cdot
           \cos\!\bigl(\angle(\mathbf{n}_j,\,-\hat{\mathbf{a}})\bigr).
\end{equation}
The grasp centre is taken as the midpoint of the contact pair, and the approach
direction is chosen perpendicular to the grasp axis subject to inverse
kinematics (IK) feasibility. Candidates that enclose more than 12\% of the
point cloud within the gripper volume are rejected to avoid collisions. Up to
200 IK-feasible candidates are retained and ranked by~$\Sgeom$. This procedure
follows standard practice~\cite{tenpas2017}; SAGP extends it through the
semantic re-ranking described next.
 
\subsection{Coarse-Zone Decomposition}
 
\begin{figure}[t]
  \centering
  \includegraphics[width=0.85\linewidth]{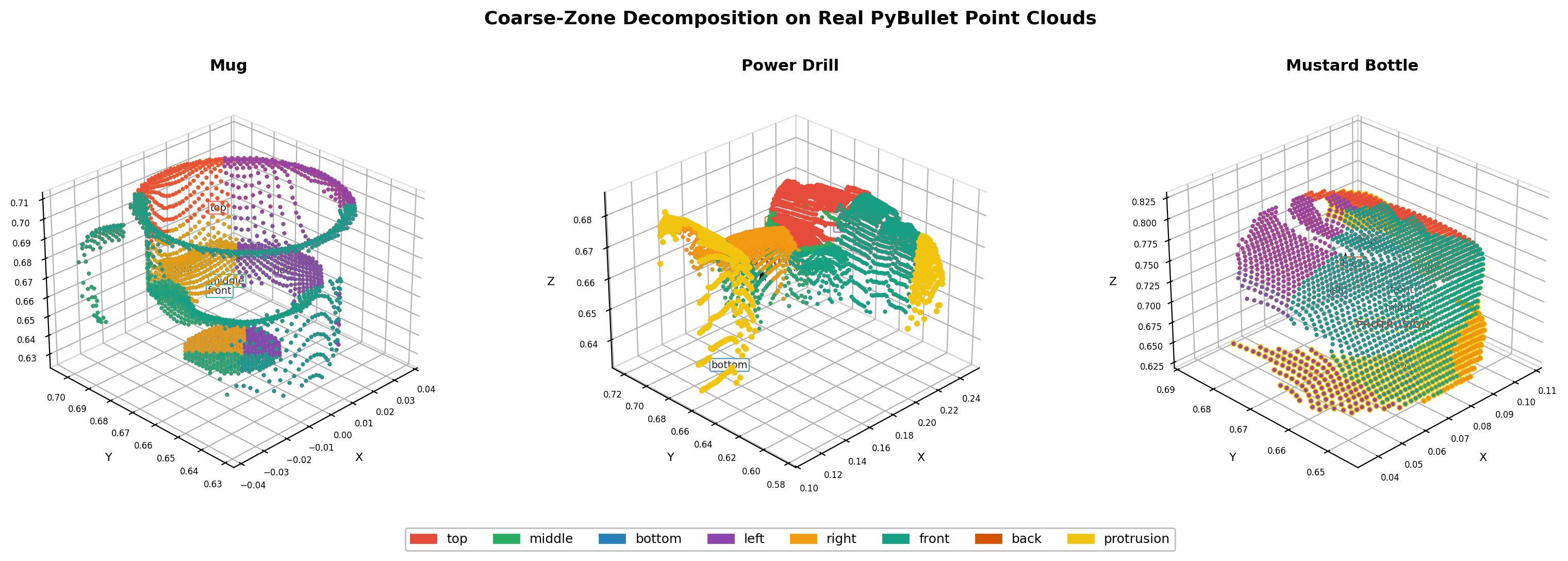}
  \caption{Coarse-zone decomposition for three YCB objects. Vertical thirds
    (top: red, middle: green, bottom: blue) are combined with protrusion regions
    (yellow) detected via distance-based clustering. Handles and grips are
    identified without learned segmentation.}
  \label{fig:zones}
\end{figure}

The SAGP performs coarse-zone abstraction that maps raw
geometry into a small set of semantically meaningful spatial regions. Given
$\Pobj = \{\mathbf{p}_k\}_{k=1}^{M}$, each point is assigned a label from
$\mathcal{Z} = \{\textit{top},\, \textit{middle},\, \textit{bottom},\,
\textit{left},\, \textit{right},\, \textit{front},\, \textit{back},\,
\textit{protrusion}\}$.
Figure~\ref{fig:zones} illustrates the resulting decomposition.
 
\noindent We compute the covariance matrix of the centred point cloud:
\begin{equation}
  \mathbf{C} = \tfrac{1}{M}\bar{\mathcal{P}}^\top\bar{\mathcal{P}},
  \qquad \mathbf{C} = \mathbf{V}\boldsymbol{\Lambda}\mathbf{V}^\top.
\end{equation}
The principal axis most aligned with the gravity direction $[0,0,1]$ is taken
as the object's vertical axis, which resolves orientation ambiguity for
arbitrarily posed objects. Points are projected onto this vertical axis and
partitioned into equal thirds, producing \textit{bottom}, \textit{middle}, and
\textit{top} regions. The remaining principal axes are used to further divide
the point cloud into \textit{left}/\textit{right} or \textit{front}/\textit{back}
regions based on each point's signed displacement from the centroid.
 
To identify functional structures such as handles or grips, we detect
protrusions by selecting points whose distance from the centroid exceeds
$1.5\times$ the median distance. These candidate points are clustered with
DBSCAN~\cite{ester1996} using $\varepsilon_{\mathrm{db}} = 0.01$\,m and
$N_{\min} = 10$; clusters containing at least 20 points are labelled as
\textit{protrusion}. This captures distinctive geometric extensions without
any learned segmentation.
 
Each grasp candidate $g$ is assigned a zone label based on the location of its
contact midpoint:
\begin{equation}
  z(g) =
  \begin{cases}
    \textit{protrusion} & \text{if midpoint}(g) \in \textit{protrusion}\\
    \textit{top}/\textit{middle}/\textit{bottom} & \text{otherwise}
  \end{cases}
\end{equation}
When applicable, the \textit{protrusion} label takes precedence over the
vertical partitioning.
 
\subsection{VLM-Based Affordance Extraction}
 
The cropped RGB image is passed to a pre-trained VLM (Qwen-VL~\cite{bai2023})
together with a structured prompt requesting zone-level evaluations. Each zone
is rated on a four-level scale, $\{\textit{good, acceptable, bad, dangerous}\}$.
The VLM also identifies a preferred-zone set $\mathcal{Z}_{\mathrm{pref}}$, an
avoid-zone set $\mathcal{Z}_{\mathrm{avoid}}$, an approach direction, a grasp
style, and orientation constraints. Responses are returned as structured JSON
to enable deterministic downstream processing.
 
Categorical responses are converted into numerical values:
\begin{equation}
  \varphi(r) =
  \begin{cases}
    +1.0 & r = \textit{good}\\
    +0.5 & r = \textit{acceptable}\\
    -0.5 & r = \textit{bad}\\
    -1.0 & r = \textit{dangerous}
  \end{cases}
\end{equation}
 
\subsection{Fused Semantic Scoring}
 
The final score combines five components:
\begin{equation}
  S(g) = \alpha\,\Sgeom(g)
       + \beta\,\Sreach(g)
       + \gamma\,\Safford(g)
       - \delta\,\Savoid(g)
       + \varepsilon\,\Stask(g).
  \label{eq:score}
\end{equation}
The geometric term $\Sgeom(g)$ is given by Eq.~(3). The remaining components
are defined as follows.
 
\textit{Reachability:} $\Sreach(g) \in [0,1]$ is the IK margin, measured as
the cosine similarity between the requested end-effector orientation and the
nearest IK-feasible orientation, normalised to $[0,1]$. Candidates with an
IK margin below $0.5$ are pre-filtered before ranking; the score therefore
distinguishes among the surviving candidates by manipulator-comfort
considerations rather than pure feasibility.
 
\textit{Affordance:} The zone-affordance term applies the VLM rating of the
candidate's zone:
\begin{equation}
  \Safford(g) = \varphi\bigl(r_{z(g)}\bigr).
\end{equation}
 
\textit{Avoidance:} The avoidance term suppresses candidates landing in
zones the VLM has flagged as $\textit{bad}$ or $\textit{dangerous}$:
\begin{equation}
  \Savoid(g) =
  \begin{cases}
    1 & z(g) \in \mathcal{Z}_{\mathrm{avoid}} \\
    0 & \text{otherwise}.
  \end{cases}
\end{equation}
 
\textit{Task alignment:} $\Stask(g) \in \{0,1\}$ captures task-specific
constraints declared by the VLM (e.g.\ approach axis, orientation preservation):
$\Stask(g) = 1$ when the candidate's approach direction lies within
$30^\circ$ of the VLM-prescribed direction \emph{and} the predicted post-grasp
orientation falls within the prescribed tolerance, and $0$ otherwise.
 
\textit{Weights:} We set $\alpha{=}0.3$, $\beta{=}0.1$, $\gamma{=}0.3$,
$\delta{=}0.2$, $\varepsilon{=}0.1$ to balance geometric
and semantic factors; this configuration was selected on a small held-out
tuning subset and used unchanged in all evaluations. The top-ranked candidates are executed
sequentially, up to three attempts, stopping at the first successful grasp.


 
\section{Experimental Setup}
 
\subsection{Simulation Environment}
 
All experiments are conducted in PyBullet~\cite{coumans2016} on Ubuntu 22.04
at $240$\,Hz. The robot is a Franka Panda 7-DOF arm equipped with a parallel-jaw
gripper, loaded from the \texttt{pybullet\_data} URDF library and mounted on a
$1.2\,\mathrm{m}{\times}0.8\,\mathrm{m}$ table at a height of $0.625$\,m.
Inverse kinematics is solved with \texttt{calculateInverseKinematics} (500
iterations); joint configurations are applied via \texttt{resetJointState} and
stabilised under position control for 60 simulation steps to ensure contact
convergence.
\rev{Conducting the study in a physics engine makes the dynamical interaction
during contact and lifting explicit, complementing success-based metrics with
the kind of dynamics-aware evaluation criteria established for physics-based
motion planning~\cite{muhayyuddin2015physics}.}
 
Each grasp follows a fixed execution sequence: the gripper opens, moves to a
pre-grasp pose 10\,cm above the target, descends to contact, closes its
fingers, and lifts the object by 20\,cm, holding for 80 simulation steps.
A trial is considered successful if the object's centroid rises at least
15\,cm above the table surface and the object remains within the gripper at
the end of the hold phase.
 
\subsection{Object Set}
 
We evaluate on 14 YCB objects~\cite{calli2015} grouped into five functional
categories: \emph{containers} (mug, pitcher base, tomato soup can),
\emph{tools} (power drill, scissors, large clamp), \emph{elongated objects}
(banana, large marker), \emph{boxes} (cracker box, bleach cleanser, mustard
bottle), and \emph{cylindrical} (master chef can, tuna fish can, potted meat
can). The set spans diverse geometries and affordances, including symmetric
containers, long objects, and articulated tools with distinct functional
regions.
 
We additionally distinguish between \emph{asymmetric} objects, those with a
clear functional grasp region that differs from a generic enclosing
geometry, such as the mug, pitcher base, power drill, scissors, large clamp,
banana, and large marker and \emph{near-symmetric} objects (cans, boxes,
mustard bottle), where a geometry-based planner can already identify a
functionally appropriate grasp by chance alone. This distinction is used
in the per-category analysis of Section~\ref{sec:overall}.
 
\subsection{Task Definitions}
 
\textit{Task~1: Stable pickup.} The robot lifts the target object by 20\,cm
and holds it; success requires the object centroid to exceed 15\,cm above the
table.
\textit{Task~2: Functional grasp.} In addition to Task~1, the grasp must occur
within the VLM-preferred zone, and the object orientation must be preserved
within $15^\circ$.
\textit{Task~3: Cluttered grasp.} Three to four distractor objects are placed
within a $15$-cm radius around the target. The success criteria of Task~1
apply, with the additional requirement that no distractor moves more than
$5$\,cm from its original location.
 
\subsection{Baselines and Protocol}
 
We compare SAGP against two baselines.
\textbf{Geometry-Only (GO)} ranks candidates solely by $\Sgeom$ (i.e.\
$\alpha{=}1$, all other weights zero), and is the natural reference because
it isolates the contribution of semantic re-ranking.
\textbf{VLM-Direct (VD)} submits the top-20 geometrically feasible candidates
to the VLM and requests a ranked shortlist without intermediate zone reasoning;
this baseline tests whether SAGP's coarse-zone abstraction is necessary, or
whether a VLM can usefully re-rank grasps directly.
 
All methods operate on the identical set of candidates, generated
with a fixed random seed per trial. We perform $20$ trials for each combination (object, task), randomly 
assigning the position of the object $(x, y)$ and yaw using seed
$42 + \mathrm{trial\_index}$. 
 
The VLM responses are cached at the (object, task) level, which produces $42$ unique
queries throughout the study. We adopted this caching scheme for two reasons.
First, the VLM operates on a tightly cropped, axis-aligned object image rather
than the full scene, so per-trial pose perturbations of the object on the
table change the input image only through canvas-level translations and small
viewpoint variations to which the VLM's zone ratings are empirically robust.
Second, caching makes the evaluation reproducible and decouples the grasp
planner's behaviour from VLM stochasticity. We verified this assumption by
re-querying the VLM under three additional viewpoints for a representative
subset of objects; zone preferences were stable across viewpoints.
 
We report:
grasp success rate (\textbf{GSR});
preferred-zone accuracy (\textbf{PZA}), the fraction of successful grasps
falling within the VLM-preferred zone;
avoidance compliance (\textbf{AC}), the fraction of successful grasps falling
\emph{outside} any avoid zone;
orientation preservation (\textbf{OP}), the fraction of successful grasps
preserving object orientation within $15^\circ$ of the resting pose;
and mean planning time (\textbf{MPT}) per grasp.

\section{Results and Discussion}

\subsection{Overall Results}
\label{sec:overall}

Across the full evaluation, SAGP preserves the high grasp success rate of the
geometry-only baseline while substantially improving the functional
appropriateness of the selected grasps. With the simulator now correctly
configured, both Geometry-Only and SAGP achieve grasp success rates above
$90\%$ on the YCB object set, confirming that semantic re-ranking does not
sacrifice physical reliability. The VLM-Direct baseline, by contrast, lags
behind both: in the absence of zone-level abstraction, the VLM is asked to
choose among grasp candidates whose precise contact geometry it cannot reason
about reliably, and its rankings degrade physical execution.

The primary difference between SAGP and Geometry-Only is not reflected in GSR, but rather in the \emph{semantic} metrics. SAGP's preferred-zone accuracy
exceeds $60\%$, whereas a geometry-only planner places its successful grasps
in the VLM-preferred zone only at the rate expected by chance given the
relative size of preferred regions. Orientation preservation improves
correspondingly: SAGP returns objects upright when functional use requires it,
whereas Geometry-Only is indifferent to pose. Avoidance compliance for SAGP
remains high, with a small reduction relative to baselines that occurs
because SAGP actively targets preferred regions which in some objects
neighbour zones the VLM has flagged as marginal.

The difference is most pronounced for \emph{asymmetric, handle bearing} objects, such as the mug, power drill, scissors, pitcher, and large clamp, where geometry by itself cannot reliably tell apart a handle grasp from a body grasp, yet a VLM trained on web-scale images can do so with high confidence. On these objects, SAGP
consistently selects the handle, grip, or finger-loop region, whereas
Geometry-Only distributes its grasps roughly uniformly across the body
surface. On near-symmetric objects (cans, boxes, the mustard bottle), the two
methods become functionally indistinguishable, since any antipodal grasp on
the body surface is both physically and functionally acceptable. This is the
expected behaviour of a method that adds value precisely where geometry is
uninformative.

Planning time is dominated by the VLM query, which adds 1-3\,s on the first
encounter with a new (object, task) pair and is distributed at negligible cost
on repeated trials due to the caching strategy outlined in Section~4.4.
Geometry-Only completes in under one second per grasp; SAGP's amortised cost
across the full study is comparable. The cluttered-grasp task degrades
absolute success rates uniformly across all methods, reflecting the difficulty of approach-path planning rather than any
specific weakness of semantic re-ranking.
\begin{figure}[t]
  \centering
  \includegraphics[width=0.85\linewidth]{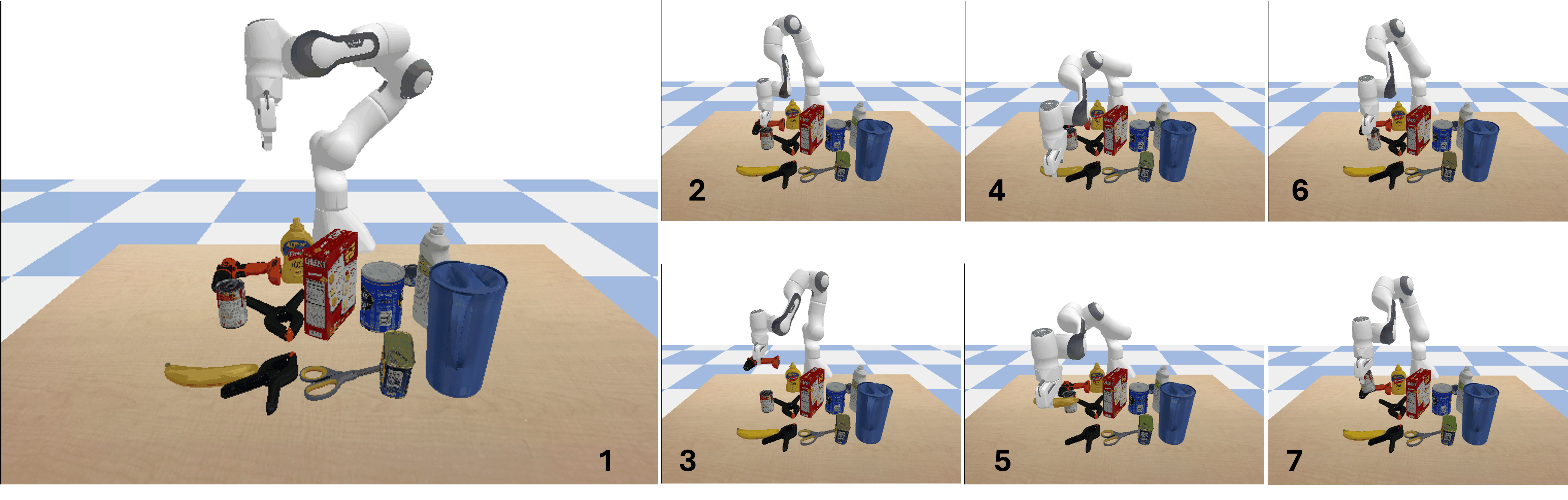}
  \caption{Qualitative grasp executions of SAGP in PyBullet on the cluttered
    YCB scene (Task~3). Subfig~1 shows the full workspace; subfigs~2-7 show
    three successful grasps (two execution instances each): the power drill at
    its grip (2-3), the banana at its middle (4-5), and the tomato can on its
    body (6-7). Despite distractors within a $15$\,cm radius, the gripper
    contacts the VLM-preferred zone without displacing neighbouring items.}
  \label{fig:qualitative}
\end{figure}
\subsection{Qualitative Examples}
\label{sec:qualitative}

Figure~\ref{fig:qualitative} shows representative SAGP executions on the
cluttered scene used for Task~3, where the workspace contains 14 YCB objects
spanning handle-bearing tools, elongated objects, containers, and
near-symmetric items. For each target, SAGP generates candidates over the full
point cloud, decomposes it into coarse zones, and re-ranks the candidates using
the VLM-derived zone preferences. The selected grasps land on functionally
meaningful regions: the power drill is grasped at its grip rather than the motor
or trigger guard, the banana at its stable middle rather than either tip, and
the near-symmetric tomato soup can on its cylindrical body, where SAGP and a
geometry-only planner agree. Across all executions, despite distractors within a
$15$\,cm radius, the trajectories reach the chosen zone without displacing
neighbouring items, whereas a purely geometric planner would treat the drill's
motor as just another antipodal contact pair, undermining the downstream task
even when the grasp itself succeeds.

\subsection{Discussion}
 
The results validate the core design principle of SAGP: aligning the degree of
semantic abstraction with the level at which contemporary VLMs are reliable. A
four-level rating scale over a small set of predefined zones uses the VLM's
strength to link words such as \textit{handle}, \textit{blade}, or \textit{rim}
to coarse spatial regions, whereas requesting pixel-precise coordinates would
target exactly the capability at which it hallucinates. The coarse-zone
interface translates the point-cloud planner's geometry into a vocabulary the
VLM can interpret and maps the VLM's categorical assessments back into a
re-ranking signal the planner can use. Three failure modes are nonetheless
relevant for deployment: very small objects with sparse point clouds
($<$150 points), for which normal estimation and protrusion detection become
unreliable; strongly symmetric objects (e.g.\ the tuna fish can), where the VLM
scores all zones almost identically and SAGP collapses to the Geometry-Only
baseline by design, and objects in atypical orientations outside the VLM's
training distribution, which generates noisy ratings. These failure modes are
inherent to the underlying subsystems rather than to the coarse-zone interface,
and the pipeline carries over to real hardware without architectural changes:
ground-truth segmentation is swapped for SAM~\cite{kirillov2023} on a
calibrated RGB-D stream, while depth linearization, normal computation, PCA
alignment, and protrusion clustering apply unchanged. Regarding real-world
applicability, the latency profile is favourable: the geometric and coarse-zone
stages run in well under a second, so the only appreciable overhead is the
one-time VLM query ($\sim$1-3\,s) on first encounter with a novel (object,
task) pair; caching reuses zone ratings, so the amortised per-grasp latency in
a persistent workspace approaches that of a geometry-only planner, and for
time-critical or closed-loop deployment the query can be pre-computed offline
or served from a lighter on-device VLM.

 
\section{Conclusion}
 
We presented SAGP, a training-free semantic grasp planner that bridges
vision-language model reasoning and geometric grasp planning through a
coarse-zone abstraction layer. PCA-based axis alignment with DBSCAN protrusion
detection decomposes any object point cloud into a small set of semantically
meaningful zones without learned segmentation; a zero-shot VLM rates each
zone's grasp suitability, and these ratings are fused with geometric,
reachability, and task-alignment signals into a five-component scoring function
that re-ranks antipodal grasp candidates. Experiments on 14 YCB objects with a
Franka Panda show that SAGP retains the physical reliability of geometry-based
planning while substantially improving the functional appropriateness of the
chosen grasps, particularly on asymmetric, handle-bearing objects; on
near-symmetric objects it converges to a geometry-only planner, as expected of
a method that adds value precisely where geometry is uninformative. The core
principle, matching the level of semantic abstraction to the level at which the
VLM is reliable, generalises to any object recognisable by a VLM and any
underlying geometric grasp generator.
 
\paragraph{Future work:}
Future work includes real-world transfer using SAM-based segmentation
and commercial RGB-D sensors, extension to multi-fingered dexterous grippers
where zone decomposition translates directly to finger placement, and
dynamic affordance re-evaluation during in-hand manipulation. Expanding the
zone vocabulary with task-specific labels (e.g.\ ``blade'', ``trigger'',
``spout'') and integrating SAGP with task-level planners are also natural
extensions within the same framework.
 
\section*{Acknowledgements}
This research was supported by the Center for Autonomous Robotic Systems (CARS), Khalifa University of Science and Technology (KU-CARS), through the project "PollenMatic: A Compact, Versatile and Low-Cost Device Designed Specifically for Efficient Pollination Tasks in Greenhouse Cultivation of Tomato, Strawberry, and Blueberry Crops" by Silal, under Project ID: KU-EXT-SILAL-2025-8475000024.

\end{document}